\documentclass[letterpaper, 10pt, conference]{IEEEtran}

\IEEEoverridecommandlockouts                              

\usepackage[numbers]{natbib}
\usepackage{graphicx}
\usepackage{amsmath}
\usepackage{hyperref}
\usepackage{algpseudocode}
\usepackage{algorithm}
\usepackage[algo2e]{algorithm2e}
\usepackage{multirow}
\usepackage[normalem]{ulem}
\usepackage{booktabs}
\usepackage{gensymb}

\usepackage{amsmath,amsfonts,bm}

\def\eqref#1{equation~\ref{#1}}

\def\1{\bm{1}}

\DeclareMathAlphabet{\mathsfit}{\encodingdefault}{\sfdefault}{m}{sl}
\SetMathAlphabet{\mathsfit}{bold}{\encodingdefault}{\sfdefault}{bx}{n}

\newcommand{\E}{\mathbb{E}}

\newcommand{\norm}[1]{\left\lVert#1\right\rVert}

\newcommand{\xxnote}[3]{}
\ifx\hidenotes\undefined
  \renewcommand{\xxnote}[3]{\color{#2}{#1: #3}}
\fi

\hypersetup{
    colorlinks=true,
    linkcolor=blue,
    filecolor=magenta,      
    urlcolor=blue,
}

\usepackage[font={small}]{caption}
\renewcommand\thefigure{\arabic{figure}}
\title{\LARGE \bf
Learning Predictive Representations for Deformable Objects \\Using Contrastive Estimation
}

\author{Wilson Yan$^{1,\dagger}$, Ashwin Vangipuram$^{1}$, Pieter Abbeel$^{1}$, and Lerrel Pinto$^{1,2}$
\thanks{
$^{1}$Department of EECS, University of California, Berkeley.}
\thanks{
$^{2}$Department of Computer Science, New York University.}
\thanks{
$^{\dagger}$Correspondence to \texttt{wilson1.yan@berkeley.edu}}
}

\begin{document}

\makeatletter
\let\@oldmaketitle\@maketitle%
\renewcommand{\@maketitle}{\@oldmaketitle%
    \centering
    \includegraphics[width=\linewidth]{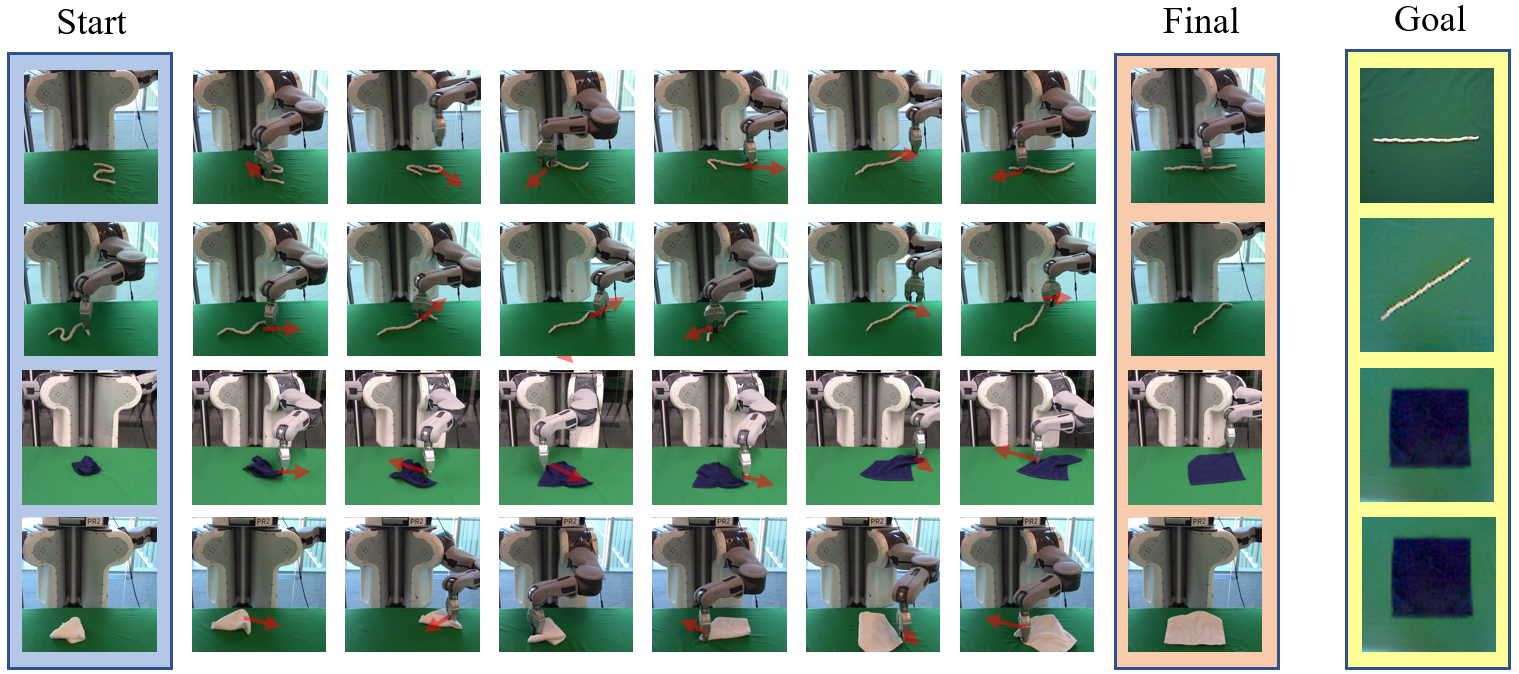}
    \captionof{figure}{Example trajectories using our contrastive forward model for rope and cloth manipulation. The top two rows show rope manipulation from different start states to different goal states, while the bottom two rows show cloth manipulation using different colored cloths. Note that in the last row, the robot is manipulating a white cloth, but are method is able to still use a blue cloth as the goal image.}
    \label{fig:intro}
}
\makeatother

\maketitle

\thispagestyle{empty}
\pagestyle{empty}

\begin{abstract}
Using visual model-based learning for deformable object manipulation is challenging due to difficulties in learning plannable visual representations along with complex dynamic models. In this work, we propose a new learning framework that jointly optimizes both the visual representation model and the dynamics model using contrastive estimation. Using simulation data collected by randomly perturbing deformable objects on a table, we learn latent dynamics models for these objects in an offline fashion. Then, using the learned models, we use simple model-based planning to solve challenging deformable object manipulation tasks such as spreading ropes and cloths. Experimentally, we show substantial improvements in performance over standard model-based learning techniques across our rope and cloth manipulation suite. Finally, we transfer our visual manipulation policies trained on data purely collected in simulation to a real PR2 robot through domain randomization.
\end{abstract}

\section{Introduction}
Robotic manipulation of rigid objects has received significant interest over the last few decades, from grasping novel objects in clutter~\cite{mahler2016dexnet,levine2016learning,shimoga1996robot,pinto2016supersizing,gupta2018robot} to dexterous in-hand manipulation~\cite{kumar2016optimal,andrychowicz2018learning,yousef2011tactile}. However, the objects we interact within our daily lives are not always rigid. From putting on clothes to packing a shopping bag, we constantly need to manipulate objects that deform. Even seemingly rigid objects like metal wires significantly deform during everyday interactions. As a result, there has been a growing interest in algorithms that can tackle deformable object manipulation ~\cite{wada2001robust,henrich2012robot,schulman2013generalization,schulman2013tracking,seita2018deep,wu2019learning,seita2019deep,maitin2010cloth,stria2014garment}.

Deformable object manipulation presents two key challenges for robots. First, unlike rigid objects, there is no direct representation of the state. Consider the manipulation problem in Figure~\ref{fig:intro}, where the robot needs to straighten a rope from a start configuration to any goal configuration. How does one track the shape of the rope? This lack of a canonical state often limits representations to discrete approximations~\cite{berenson2013manipulation}. Second, the dynamics of deformable objects are complex and non-linear~\cite{essahbi2012soft}. Due to microscopic interactions within the object, even simple objects can exhibit complex and unpredictable behavior~\cite{pieranski2001tight}, which makes modeling and performing traditional task and motion planning with such deformable objects difficult.

One class of techniques that circumvents the challenges in state estimation and dynamics modeling is image-based model-free learning~\cite{haarnoja2018soft,schulman2015trust,lillicrap2015continuous}. For instance, \citet{matas2018sim,seita2019deep,wu2019learning} use model-free methods in simulation for several difficult cloth manipulation tasks. However, without expert demonstrations, model-free learning is notoriously inefficient~\cite{duan2016benchmarking}, and often needs millions of samples to learn from. This challenge is further exacerbated in the multi-task learning framework, where the robot needs to learn to reach multiple goals. 

Model-based techniques, on the other hand, have shown promise in sample-efficient learning~\cite{williams2017information,camacho2013model,nagabandi2018neural}. However, using such model-based learning techniques for deformable objects necessitates tackling the challenges of state representation and dynamics modeling head-on. So how does one learn models given high-dimensional observations and complex underlying dynamics? Some approaches take a direct approach to learning complex dynamics models through pixel-space~\cite{kaiser2019model,ebert2018visual}. Another approach, by \citet{agarwal2016,nair2017combining}, learns forward dynamics models in conjunction with inverse dynamic models for manipulating deformable objects. However, during robotic execution, only the inverse model is used. Other model-based approaches such as \citet{wang2019learning} train Causal InfoGANs~\cite{kurutach2018learning,chen2016infogan} to both extract visual representations and forward models, and use the learned forward models for planning. However, these techniques are not robust due to training instabilities associated with GANs~\cite{srivastava2017veegan}.

In this paper, we introduce a new visual model-based framework that uses contrastive optimization to jointly learn both the underlying visual latent representations and the dynamics models for deformable objects. We hypothesize that using contrastive methods for model-based learning achieves better generalization and latent space structure do to its inherent information maximization objective. We re-frame the objective introduced in contrastive predictive coding~\cite{oord2018representation} to allow for learning effective model dynamics and latent representations. Once the latent models for representations and dynamics are learned across offline random interactions, we use standard model predictive control (MPC) with one-step predictions to manipulate deformable objects to desired visual goal configurations. Given this controller, we empirically demonstrate substantial improvements over standard model-based learning approaches across multi-goal rope and cloth spreading manipulation tasks. Videos of our real robot runs and reference code can be found on the project website: \url{https://sites.google.com/view/contrastive-predictive-model}.

In summary, we present three key contributions in this paper: (a) We propose a contrastive predictive modeling approach to model learning that is compatible with model predictive control.
To our knowledge, this is the first use of contrastive estimation for model-based learning. 
(b) We demonstrate substantial improvements in multi-task deformable object manipulation over other model learning approaches. (c) We show the applicability of our method to real robot rope and cloth manipulation tasks by using sim-to-real transfer without additional real-world training data.

\section{Related Work}

\subsection{Deformable Object Manipulation}
There has been a substantial amount of prior work in the area of robotic manipulation of deformable objects. A detailed survey of past work can be found in~\citet{khalil2010dexterous,henrich2012robot}.

A standard approach to tackling deformable object manipulation is to use deformable object simulations with planning methods~\cite{jimenez2012survey}. Past work in this domain has focused on simple linear deformable objects~\cite{saha2007manipulation,wakamatsu2006knotting,moll2006path}, creating better simulations~\cite{rodriguez2006obstacle}, and faster planning~\cite{frank2011efficient}. However, the large number of states for deformable objects makes it difficult to plan correctly while being computationally efficient.

Instead of directly planning on the full dynamics, some prior research has focused on planning on simpler approximations, by using local controllers to handle the actual complex dynamics. One approach to using local controllers is model-based servoing~\cite{smolen2009deformation,wada2001robust}, where the end-effector is controlled to a goal location instead of explicit planning. However, since the controller is optimized over simple dynamics, it often gets stuck in local minima with more complex dynamics~\cite{mcconachie2017interleaving}. To solve this, several works~\cite{berenson2013manipulation,mcconachie2018estimating} have proposed Jacobian controllers that do not need explicit models, while \cite{jia2018learning,hu2018three} have proposed learning-based techniques for servoing. We note that our proposed work on learning latent dynamics models is compatible with several of these model-based optimization techniques. 

\subsection{Contrastive Prediction}
Learning good representations remains a difficult challenge in deformable object manipulation. There has been a large amount of prior work on contrastive predictive methods to learn better representations of data. Word2Vec~\cite{mikolov2013distributed} optimizes a contrastive loss to demonstrate semantic and syntactic structure in the learned latent space for words. \citet{oord2018representation} shows that it is possible to learn high-level representations of images, video, and speech data by employing a large number of negative samples. \citet{tian2019contrastive} learns high-level representations by encouraging different views of scenes to be embedded close to one another, and further from others through a similarly framed contrastive loss. Recently, SimCLR~\cite{chen2020simple}, another contrastive learning framework, achieved state-of-the-art results in self-supervised learning representations, bridging the gap with supervised learning.

\section{Contrastive Forward Modeling (CFM)}
\begin{figure}
    \centering
    \includegraphics[scale=.5]{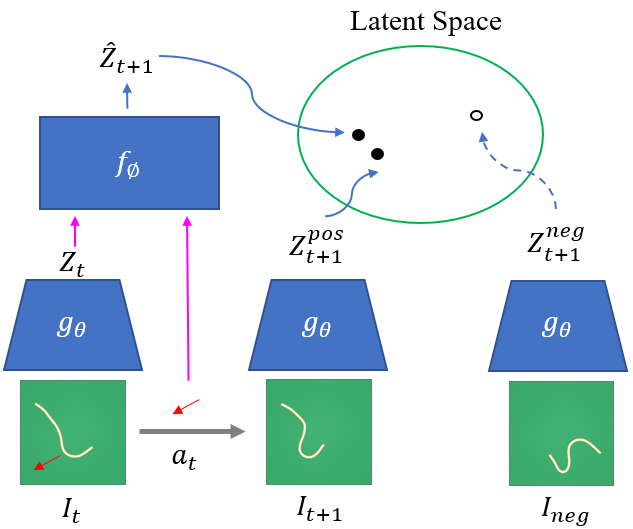}
    \caption{Overview of our contrastive forward model. Training data consists of (image, next image, action) tuples and we learn the encoder and forward model jointly. The contrastive loss objective brings the positive embedding pairs closer together and the negative embeddings further away.}
    \label{fig:cpc_model}
\end{figure}

In this section, we describe our proposed framework for learning deformable object manipulation: Contrastive Forward Modeling (CFM). We begin by discussing formalism for predictive modeling and contrastive learning. Following that, we discuss our method for learning contrastive predictive models. See Figure~\ref{fig:cpc_model} for an overview of our training scheme.

\subsection{Dynamic Predictive Models} 
For our problem setting, we consider a fully observable environment with observations $o \in \mathcal{O}$, actions $a \in \mathcal{A}$, and deterministic transition dynamics $f(o_t, a_t) = o_{t+1}$. We would like to learn a predictive model $\hat{f}(o_t, a_t) \approx o_{t+1}$ to approximate the observation of the next timestep. This can be done by directly learning a visual model through pixel space with regression over observation-action-observation tuples~\cite{finn2017deep,kaiser2019model}. Once we have successfully learned a predictive model, it is natural to use it for planning to reach different desired goal states, for example, different configurations of a rope or cloth. However, planning directly through pixel space can be difficult, as pixel-value comparisons between images usually do not necessarily correlate well with their true distances. For example, consider an environment with a ball, where the task is to learn a policy that pushes the ball to the center. If the ball is far from the center, then all predicted next actions using a visual forward model would be equidistant from the goal ball-in-center image when comparing pixels values since there would be no image overlap. Therefore, we consider the framework of planning with in a learned latent space by encoding observations. We learn an encoder $g_\theta(o_t) = z_t$ to embed our observations into a latent space, coupled with a predictive model in latent space between $z$'s, where our learned predictive model is now formulated as $\hat{f}(z_t, a_t) \approx z_{t+1}$. In this work, we propose to learn the latent space using a contrastive learning method.

\subsection{Contrastive Models}
In our contrastive learning framework, we jointly learn an encoder $g_\theta(o_t) = z_t$ and a forward model $f_\phi(z_t, a_t) \approx z_{t+1}$. We use the InfoNCE contrastive loss described by \citet{oord2018representation}. 

\begin{equation}
\mathcal{L} = -\E_{\mathcal{D}}\left[\log{\frac{h(\hat{z}_{t+1}, z_{t+1})}{\sum_{i=1}^k h(\hat{z}_{t+1}, \tilde{z}_i)}}\right]
    \label{eq:cpc}
\end{equation}

\noindent where $h$ is some similarity function between the computed embeddings from the encoder. The $\tilde{z}_i$ represents negative samples, which are incorrect embeddings of the next state, and we use $k$ such negative samples in our loss. The motivation behind this learning objective lies with maximizing mutual information between the predicted encodings and their respective positive samples. Within the embedding space, this results in the positive sample pairs being aligned together but the negative samples pushed further apart, as seen in Figure~\ref{fig:cpc_model}. Since we are jointly learning a forward model that seeks to minimize $\norm{f_\phi(z_t, a_t) - z_{t+1}}^2$, we use the similarity function:

\begin{equation}
h(z_1, z_2) = \exp({-\norm{z_1-z_2}^2})
\label{eq:sim}
\end{equation}

\noindent where the norm is a $\ell_2$-norm. After learning the encoder and dynamics model, we plan using a simple version of Model Predictive Control (MPC), where we sample several actions, run them through the forward model from the current $z_t$, and choose the action $a_t$ that produces $\hat{z}_{t+1}$ closest (in $\ell_2$-distance) to the goal embedding.

\section{Experimental Evaluations}
In this section, we experimentally evaluate our method in various rope and cloth manipulation settings, both in simulation and in the real world. Our experiments seek to address the following questions:
\begin{itemize}
    \item Do contrastive learning methods learn better latent spaces and forward models for planning in deformable object manipulation tasks?
    \item What aspects of our contrastive learning methods contribute the most to performance?
    \item Can we successfully manipulate deformable objects on a real-world robot?
\end{itemize}

\subsection{Environments and Tasks}
To simulate deformable objects such as cloth and rope, we used the Deep Mind Control~\cite{tassa2018deepmind} platform with MuJoCo 2.0~\cite{todorov2012mujoco}. We use an overhead camera that renders $64 \times 64 \times 3$ RGB images as input observations for training our method. 

We design the following tasks in simulation:
\par \textbf{1. Rope}: The rope is represented by 25 geoms in simulation with a four-dimensional action space: the first $2$ are the pixel pick point on the rope, and the last $2$ are the $x, y$ delta direction to perturb the rope. At the start of each episode, the rope's state is randomly initialized by applying 120 random actions.
\par \textbf{2. Cloth}: The cloth is represented by a $9\times 9$ grid of geoms in simulation with a five-dimensional action space: the first $2$ are the pixel pick point on the cloth, and the last $3$ are the $x, y, z$ delta direction to perturb the cloth. At the start of each episode, the cloth's state is randomly initialized by applying $50$ random actions. In MuJoCo 2.0, the skin of the cloth can be changed by using images taken of a real cloth.

For both rope and cloth environments, we evaluate our method by planning to a desired goal state image and computing the sum of the pairwise geom distances between the achieved and true goal states. We observe that taking an average of 1000 trials suffices to maintain high-confidence evaluation estimates.

\begin{figure*}
    \centering
    \includegraphics[width=\textwidth]{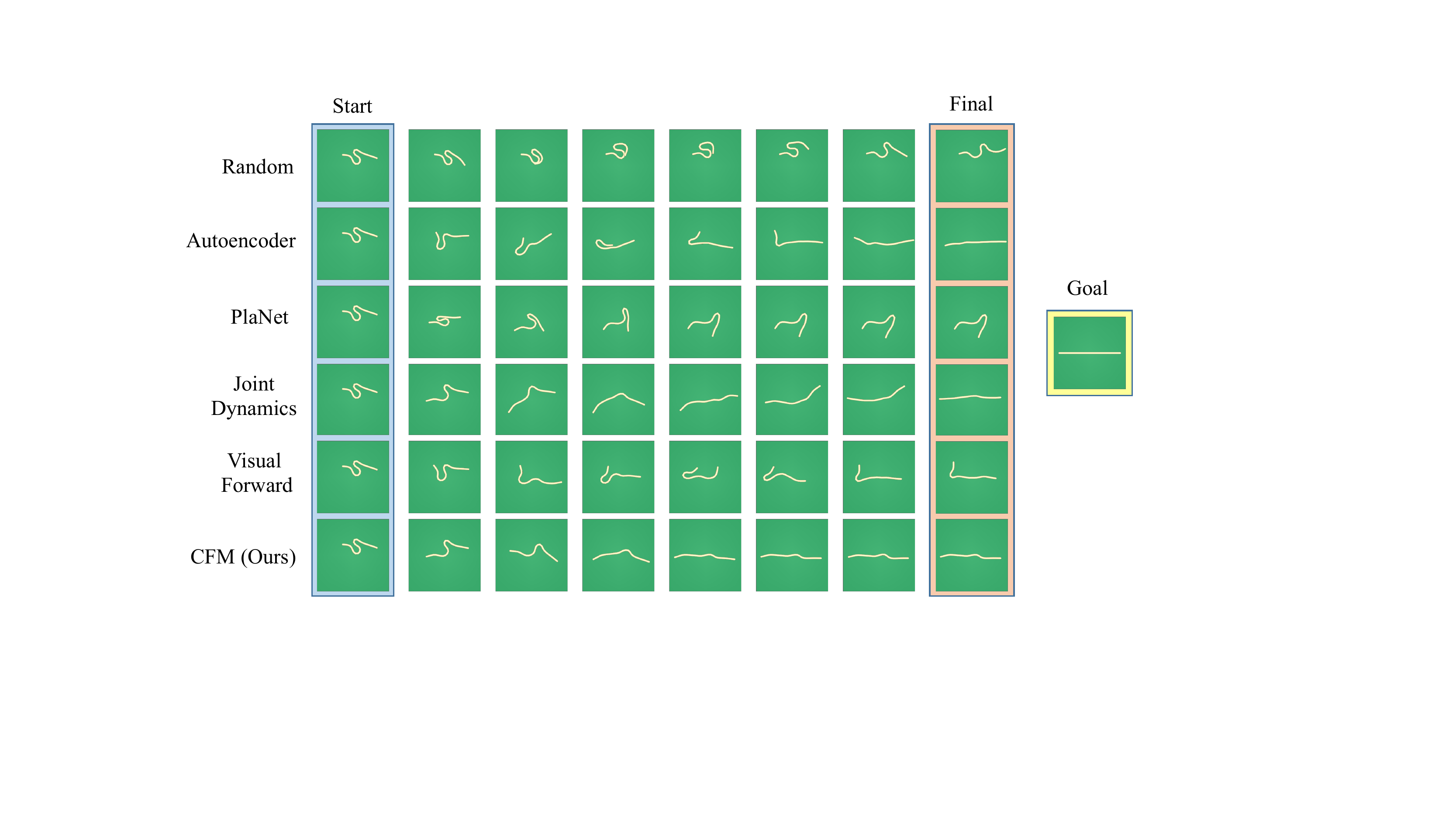}
        \caption{Trajectories for each of the baselines within the simulator, all starting from the same start state and having the same end goal of a horizontal line. Each trajectory was run for 20 actions. Note that our method (CFM) reaches the goal state significantly faster than the baselines.}
    \label{fig:sim_rope_baselines}
\end{figure*}

\begin{table*}
\caption{Quantitative comparisons between different model-based learning methods on rope and cloth manipulation tasks. The metric is the sum of pairwise geom distances between the final observation and goal state, where lower distance is more accurate.}
\centering
\begin{tabular}{c|ccccl|cl}
\toprule
                                     & \multicolumn{5}{c|}{\textbf{Rope}}                                                                                                   & \multicolumn{2}{c}{\textbf{Cloth}}                  \\
\multicolumn{1}{l|}{}                & \textbf{Horizontal}      & \textbf{Vertical}        & \textbf{45\degree}        & \textbf{135\degree}       & \textbf{Random}          & \textbf{Flat}            & \textbf{Random}          \\ \midrule
\textbf{Random Policy}               & $4.75$                   & $4.93$                   & $4.80$                   & $4.87$                   & $5.73$                   & $7.98$                   & $10.12$                  \\
\textbf{Autoencoder}                 & $1.72 \pm 0.31$          & $3.24 \pm 1.28$          & $\mathbf{2.11 \pm 0.51}$ & $2.49 \pm 0.64$          & $4.308 \pm 1.16$        & $3.24 \pm 0.29$          & $4.82 \pm 0.0$           \\
\textbf{PlaNet}                      & $1.81 \pm 0.13$          & $3.36 \pm 0.78$          & $2.31 \pm 0.72$          & $2.38 \pm 0.20$          & $3.037 \pm 0.24$        & $4.12 \pm 0.21$          & $5.06 \pm 0.02$          \\
\textbf{Joint Dynamics Model} & $2.13 \pm 0.66$          & $4.33 \pm 0.85$          & $3.88 \pm 0.95$          & $4.02 \pm 0.85$          & $1.78 \pm 0.09$        & $4.24 \pm 0.06$          & $4.70 \pm 0.03$          \\
\textbf{Visual Forward Model}        & $2.09 \pm 0.13$          & $\mathbf{2.65 \pm 0.27}$ & $2.55 \pm 0.34$          & $2.27 \pm 0.17$          & $4.77 \pm 0.18$        & $\mathbf{2.20 \pm 0.05}$ & $4.65 \pm 0.10$          \\
\textbf{CFM (Ours)}                  & $\mathbf{0.58 \pm 0.09}$ & $3.08 \pm 1.19$          & $2.29 \pm 1.42$          & $\mathbf{2.24 \pm 0.90}$ & $\mathbf{1.52 \pm 0.10}$ & $2.69 \pm 0.25$          & $\mathbf{3.97 \pm 0.16}$
\end{tabular}

\centering
\begin{tabular}{c|ccccl|cl}
\toprule
                                     & \multicolumn{5}{c|}{\textbf{Rope (With DR)}}                                                                                                             & \multicolumn{2}{c}{\textbf{Cloth (With DR)}}                      \\
\multicolumn{1}{l|}{}                & \textbf{Horizontal}        & \textbf{Vertical}          & \textbf{45\degree}           & \textbf{135\degree}         & \textbf{Random}            & \textbf{Flat}              & \textbf{Random}            \\ \midrule
\textbf{Random Policy}               & $4.75$                    & $4.93$                    & $4.80$                    & $4.87$                    & $5.73$                    & $7.975$                    & $10.12$                   \\
\textbf{Autoencoder}                 & $3.29 \pm 1.08$          & $3.70 \pm 1.47$          & $3.19 \pm 1.14$          & $3.30 \pm 1.14$          & $4.31 \pm 1.16$          & $6.26 \pm 1.23$          & $7.08 \pm 2.22$          \\
\textbf{PlaNet}                      & $2.35 \pm 0.56$          & $4.06 \pm 1.84$          & $3.73 \pm 1.66$          & $3.58 \pm 1.46$          & $3.04 \pm 0.24$          & $8.74 \pm 0.55$          & $10.10 \pm 1.56$         \\
\textbf{Joint Dynamics Model} & $1.01 \pm 0.40$          & $2.29 \pm 0.10$          & $1.35 \pm 0.59$          & $1.82 \pm 0.50$          & $1.78 \pm 0.09$          & $4.17 \pm 0.17$          & $4.64 \pm 0.20$          \\
\textbf{Visual Forward Model}        & $3.05 \pm 0.45$          & $5.65 \pm 0.37$          & $5.37 \pm 0.90$          & $5.11 \pm 1.04$          & $4.77 \pm 0.18$          & $6.64 \pm 0.66$          & $6.07 \pm 0.52$          \\
\textbf{CFM (Ours)}                  & $\mathbf{0.88 \pm 0.21}$ & $\mathbf{1.20 \pm 0.07}$ & $\mathbf{0.99 \pm 0.07}$ & $\mathbf{0.99 \pm 0.17}$ & $\mathbf{1.38 \pm 0.03}$ & $\mathbf{3.99 \pm 0.15}$ & $\mathbf{4.40 \pm 0.06}$
\end{tabular}
\label{table:main_results}
\end{table*}

\subsection{Data Collection}
Since collecting real-world data on robots is expensive, our method seeks to address this problem by collecting randomly perturbed rope and cloth data in simulation. Using random perturbations allows for a diverse set of deformable objects and interactions for learning the latent space and dynamics model. We collect 4000 trajectories of length 50 for rope (200k samples), and 8000 trajectories of length 50 for cloth (400k samples). 

\subsection{Baselines}
To show the substantial improvements of our model over prior methods, we compare our method against several baselines: a random policy, a visual forward model, an autoencoder trained jointly with a latent dynamics model, PlaNet~\cite{hafner2018learning}, and a joint dynamics model~\cite{agarwal2016}. In order to ensure that pick points are always on the rope or cloth, we constrain our pick points using a binary segmentation of the observation image computed by RGB thresholding. During plannig, all methods use MPC with one-step prediction.

\begin{itemize}
\item \textbf{Random Policy}: We sample pick actions uniformly over the binary segmentation, and place actions are sampled uniformly random in a unit square centered around the pick location.
\item \textbf{Visual Forward Model}: We train a forward model similar to~\citet{kaiser2019model} to perform modeling and planning purely through pixel space.
\item \textbf{Autoencoder}: We learn a simple latent space model by jointly training a classical autoencoder with a forward dynamics model. The autoencoder learns to minimize the $\ell_2$-distance between reconstructed and actual images~\cite{lange2010deep}.
\item \textbf{PlaNet}: We train PlaNet~\cite{hafner2018learning}, a stochastic variant of an autoencoder, as another latent space model. PlaNet models a sequential VAE and optimizes a temporal variational lower bound.
\item \textbf{Joint Dynamics Model}: We jointly learn a forward and inverse model following \citet{agarwal2016}.
\end{itemize}

For consistency across all latent space models, we use a latent size of $8$ for both the rope and the cloth environments. For all methods, we sample $100$ possible one-step trajectories when performing closed-loop planning. See Figure~\ref{fig:sim_rope_baselines} for example trajectories from each baseline in comparison to our method.

\subsection{Training Details}
We used the same encoder architectures for all models. The encoder architecture is a series of 6 2D convolutions of kernel sizes $[3, 4, 3, 4, 4, 4]$, strides $[1, 2, 1, 2, 2, 2]$, and filter sizes $[64, 64, 64, 128, 256, 256]$ respectively. We add Leaky ReLU~\cite{maas2013rectifier} activation in between each convolutional layer. Finally, the output is flattened and fed into a fully connected layer to produce the latent $z$. The forward model is a multi-layer perceptron~(MLP) with two hidden layers of size $32$ which outputs the parameters for a linear transformation on $z_t$. Specifically for our method (CFM), we use the other batch elements as our negative samples for a total of $127$ negative samples per positive pair. For PlaNet, following \citet{hafner2018learning}, the decoder architecture is a dense layer followed by $4$ transposed convolutions with kernel size $4$ and stride $2$ to upscale to the size of the $64\times 64$ image. The visual forward models follow the same convolutional encoder and decoder architectures as the previous model, with action conditioning implemented in a similar way to~\citet{kaiser2019model} where actions are processed by a separate dense layer at each resolution, multiplied channel-wise, and broadcasted spatially. The images and actions were centered and scaled to the range of $[-1, 1]$. We trained all models with batch size 128, learning rate $10^{-3}$, and an Adam optimizer~\cite{kingma2014adam} for $30$ epochs. Each model was trained on a single NVIDIA TitanX GPU, taking roughly $1-2$ hours. All of our simulated environments, evaluation metrics, and training code will be publicly released.

\subsection{Does Using Contrastive Models Improve Performance?}
In this section, we compare the results of using our method with those of our baselines, analyzing the advantages and benefits that contrastive models bring over prior methods. Consider a naive baseline where we replace the InfoNCE loss with an MSE loss. This is equivalent to jointly fitting an encoder and dynamics model that minimizes $\norm{f_\phi(g_\theta(o_t), a_t) - z_{t+1}}^2$. We can see that the optimal solution is for the encoder to encode all observations to a constant vector to achieve zero loss. To prevent this form of a degenerate solution, we are required to regularize our latent space in some way. Both prior methods and contrastive learning do this in different ways so we analyzed which methods performed better over others. Table~\ref{table:main_results} shows the quantitative results comparing our method against baselines in different rope and cloth environments, with and without domain randomization for robot transfer. Note that our method does better on all randomly sampled goals with and without domain randomization, indicating stronger generalization in latent spaces for planning. Figure~\ref{fig:sim_rope_baselines} shows example simulator trajectories for each baseline. Each trajectory has the same starting location, same goal image, and was run for 20 actions. 

An autoencoder regularizes its latent space by requiring additionally training a decoder to learn to reconstruct $o_t$ from $z_t$. The model does well in some scenarios, such as a $45\degree$ diagonal, but performs poorly when domain randomization is introduced to allow for transfer to a real robot. This is most likely because the autoencoder is optimized to have pixel-level perfect reconstructions, so features such as lighting and color must be encoded in the latent space even when they are not needed for the task. PlaNet behaves similarly to the autoencoder, as it is also a form of a stochastic autoencoder. It performs reasonably competitive with our method but again fails when domain randomization is introduced.

The joint dynamics model regularizes its latent space by jointly learning an inverse model with the forward model. The joint model performed the best across all the baselines when moving to domain randomized data. However, our method still outperforms the joint model for every task.

The visual forward model is the only method that plans in pixel space. It generally performs poorly for tasks with objects with low area coverage, such as the different rope goal orientations, but does better than our method on the cloth flattening task. However, since the forward model operates purely in pixel space, it unsurprisingly suffers from a sharp degradation in performance when introducing domain randomization. As such, it generalizes poorly to the real robot setting.

\begin{table}[H]
\caption{Ablation experiments on forward model architecture and similarity functions, using the same evaluation metric as Table~\ref{table:main_results} (lower is better).}

\centering
\begin{tabular}{@{}lll@{}}
\toprule
                              & \textbf{Rope}   & \textbf{Cloth}  \\ \midrule
\textbf{Linear}               & $2.30 \pm 0.31$ & $4.10 \pm 0.03$ \\
\textbf{MLP}                  & $1.56 \pm 0.10$ & $3.95 \pm 0.10$ \\
\textbf{Log-bilinear Similarity} & $3.25 \pm 0.43$ & $4.16 \pm 0.15$ \\
\textbf{Ours}                 & $1.40 \pm 0.02$ & $3.81 \pm 0.09$ 
\end{tabular}
\label{table:ablation}
\end{table}

\begin{figure*}
    \centering
    \includegraphics[width=1.0\textwidth]{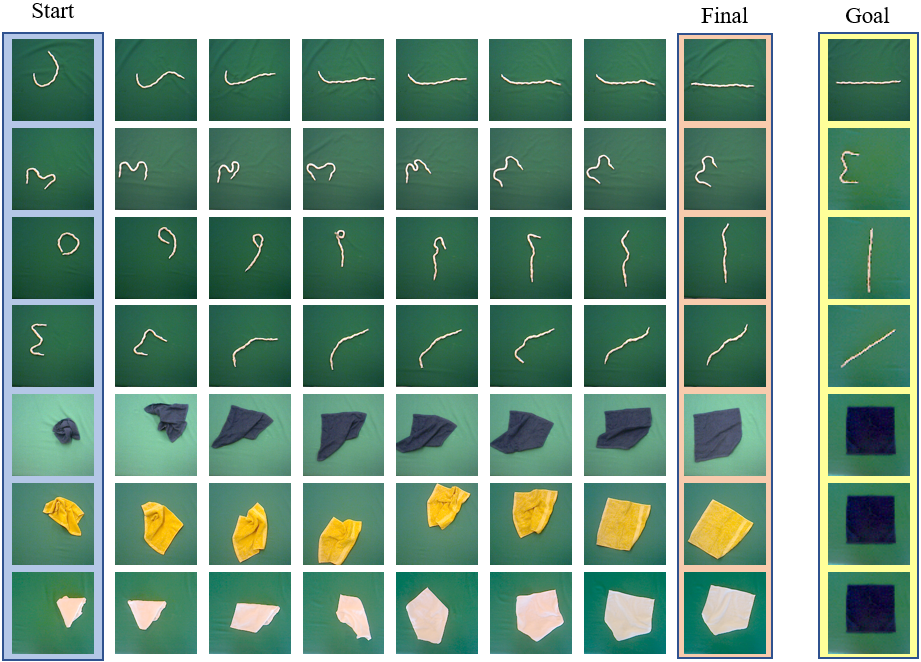}
    \caption{Each row represents one trajectory using our contrastive forward policy. The rope task uses 40 actions between the start and final states, while the cloth task uses 100 actions.}
    \label{fig:robot_traj}
\end{figure*}

\begin{table*}
\caption{The maximum intersection area in pixels between the goal image and observation images averaged over all seeds}

\begin{center}
\begin{tabular}{@{}lllllll@{}}
\toprule
Robot Experiments (Intersection in pixels) & Rope (Horizontal) & Rope (Vertical) & Rope (45\degree)       & Rope (135\degree)      & Rope (Squiggle) & Cloth (Flat)      \\ \midrule
Random Policy                                   & 6.880             & 14.727          & 13.662          & 4.266           & 0.049           & 462.513           \\
Autoencoder                                     & 5.526             & 3.334           & 3.862           & 7.499           & 3.419           & 603.927           \\
Joint Dynamic Model                     & 17.722            & 23.636          & 33.631          & 21.267          & 18.311          & 772.303           \\ \midrule
\textbf{Contrastive Forward Model (Ours)}       & \textbf{32.827}   & \textbf{36.387} & \textbf{33.891} & \textbf{38.952} & \textbf{20.711} & \textbf{1001.082} \\ \bottomrule
\end{tabular}
\end{center}
\label{table:robot_exps}
\end{table*}

\subsection{Ablations on Contrastive Models}
In this section, we perform an ablation study on our method, examining the impact of architectural designs on performance. We ablate over two aspects of our method: the forward model architecture, and the contrastive similarity function. For the forward model, our method uses a Multi-Layer Perceptron~(MLP) that outputs the parameters of a linear function that is then applied to $z_t$. For the contrastive similarity function, our method follows Equation~\ref{eq:sim}. The quantitative results, measured as the sum of pairwise geom distances between the final and goal images, appear in Table~\ref{table:ablation}.
\subsubsection{Contrastive Similarity Functions}
We compare using our similar function with the original InfoNCE similarity function in~\citet{oord2018representation}, the log-bilinear similarity function $h(z_1, z_2) = \exp(z_1^Tz_2)$. We achieve the largest boost in performance when switching to our similarity function, as it is more in line with the minimization objective of learning a correct forward model, whereas the log-bilinear model only encourages alignment (as opposed to closeness) of embedding vectors.

\subsubsection{Forward models architectures}
We experiment with a few different forward model architectures: linear, a small MLP, and a small MLP that outputs parameters for a linear transformation. As expected, the biggest drop in performance occurs when learning the simpler linear dynamics model, and a slight drop when using an MLP for both rope and cloth tasks. This demonstrates the need for more complex models for latent forward-dynamics learning.



\begin{figure*}
    \centering
    \includegraphics[width=\textwidth]{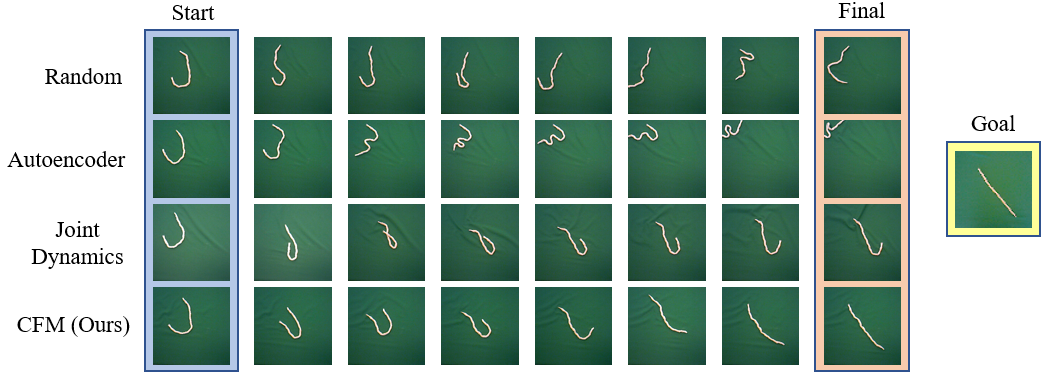}
    \caption{Rope trajectories for each baseline and our method applied on a real robot, all with the same start state, and $135^\circ$ goal rope orientation.}
    \label{fig:real_rope_baselines}
\end{figure*}

\subsection{Real Robot Experiments}
\subsubsection{Real Robot Setup}
We use a PR2 robot to perform our experiments and an overhead camera looking down on the deformable objects to get the RGB image inputs. To ensure the policy learned in the simulator transfers over to the real world, we apply domain randomization by changing the lighting, texture, friction, damping, inertia, and mass of the object during every training step within the simulator. We also use a pick and place strategy to mimic the same four-dimensional actions within the simulator. 

To compute the actions, we employ a model predictive control (MPC) approach of replanning our action at each time step based on the previous image. We segment the rope/cloth against the background to get the list of valid pick locations of the object. We then generate possible actions by uniformly sampling 100 random deltas in $[-1,1]$ combined with randomly chosen start locations. We feed these into our forward model along with the encoding of our start image to get the latent encoding for each of the next prospective states. To pick the optimal action, we find the location and delta that minimizes the Euclidean distance from these next states to our goal state and return this action to the robot. The delta from the policy is on the scale of $[-1, 1]$ for both $x$ and $y$ coordinates, and we rescale this to $[-5, 5]$ pixels. On the robot side, we use a learned linear mapping to transform from the image's pixel values to Cartesian coordinates that the robot uses. To emulate the simulator, the robot's left arm motion is to go to the start location, go down and close the gripper, move up, move to the new location, move down and open the gripper, where the height of the gripper is hard-coded to some manually tuned value. 

\subsubsection{Evaluation Metrics}
We use three baselines along with our contrastive method for real-world evaluation. The first is random actions and the others are the two policies that performed the best with domain randomization: the autoencoder and the joint dynamics model~\cite{agarwal2016}. For the rope, all the models are evaluated on five goal states: horizontal, vertical, straight line at $45^\circ$, straight line at $135^\circ$, and a squiggly rope on left. 
For the cloth, the models are evaluated on one goal state, a flat blue cloth with no rotation.
The metric we use is the intersection in pixels between the segmented final image and the segmented goal image. We prefer this instead of intersection over union (IOU) since the objects have the same shape so the union normalization is unnecessary. Additionally, the simpler intersection values provide more insight for comparisons than IOU. The models are run for 40 actions on the rope or 100 actions on the cloth, and the image after each action is stored as an observation. Among all the observations, the one with the highest intersection with the goal is chosen for each method. To account for different seeds, we use 4 starting locations for our contrastive method and 2 starting locations for the baselines, with the scores being averaged across the different start locations. For the cloth, the seed also involves different colors of cloths (blue, gold, white). 

The specific evaluation metrics are found in Table~\ref{table:robot_exps} which shows that our model performed the best for all the rope and cloth tasks. The joint dynamics model is the second best and got close results to ours on the $45^\circ$ and squiggle rope tasks. Some example trajectories from our model are seen from a forward view in Figure~\ref{fig:intro} and from an overhead view in Figure~\ref{fig:robot_traj}. Visual comparisons between our method and baseline methods on the real robot are found in Figure~\ref{fig:real_rope_baselines}. We see that our method more accurately plans towards correct goal states compared to the baselines.





\section{Conclusion}
In this paper, we propose a contrastive learning approach for predictive modeling of deformable objects. We show that contrastive learning learns stronger and more plannable latent representations compared to existing methods. Since our method only requires collecting random data in an environment, it allows for easier transfer to real robots without the need for real-world training.

\section*{Acknowledgements}
We thank AWS for computing resources. We also gratefully
acknowledge the support from Berkeley DeepDrive, NSF, and the
ONR Pecase award.


\small
\bibliographystyle{plainnat}
\bibliography{IEEEexample}

\begin{thebibliography}{59}
\providecommand{\natexlab}[1]{#1}
\providecommand{\url}[1]{\texttt{#1}}
\expandafter\ifx\csname urlstyle\endcsname\relax
  \providecommand{\doi}[1]{doi: #1}\else
  \providecommand{\doi}{doi: \begingroup \urlstyle{rm}\Url}\fi

\bibitem[Agrawal et~al.(2016)Agrawal, Nair, Abbeel, Malik, and
  Levine]{agarwal2016}
Pulkit Agrawal, Ashvin Nair, Pieter Abbeel, Jitendra Malik, and Sergey Levine.
\newblock Learning to poke by poking: Experiential learning of intuitive
  physics.
\newblock \emph{NIPS}, 2016.

\bibitem[Andrychowicz et~al.(2018)Andrychowicz, Baker, Chociej, Jozefowicz,
  McGrew, Pachocki, Petron, Plappert, Powell, Ray,
  et~al.]{andrychowicz2018learning}
Marcin Andrychowicz, Bowen Baker, Maciek Chociej, Rafal Jozefowicz, Bob McGrew,
  Jakub Pachocki, Arthur Petron, Matthias Plappert, Glenn Powell, Alex Ray,
  et~al.
\newblock Learning dexterous in-hand manipulation.
\newblock \emph{arXiv preprint}, 2018.

\bibitem[Berenson(2013)]{berenson2013manipulation}
Dmitry Berenson.
\newblock Manipulation of deformable objects without modeling and simulating
  deformation.
\newblock In \emph{IROS}, 2013.

\bibitem[Camacho and Alba(2013)]{camacho2013model}
Eduardo~F Camacho and Carlos~Bordons Alba.
\newblock \emph{Model predictive control}.
\newblock Springer Science \& Business Media, 2013.

\bibitem[Chen et~al.(2020)Chen, Kornblith, Norouzi, and Hinton]{chen2020simple}
Ting Chen, Simon Kornblith, Mohammad Norouzi, and Geoffrey Hinton.
\newblock A simple framework for contrastive learning of visual
  representations.
\newblock \emph{arXiv preprint}, 2020.

\bibitem[Chen et~al.(2016)Chen, Duan, Houthooft, Schulman, Sutskever, and
  Abbeel]{chen2016infogan}
Xi~Chen, Yan Duan, Rein Houthooft, John Schulman, Ilya Sutskever, and Pieter
  Abbeel.
\newblock Infogan: Interpretable representation learning by information
  maximizing generative adversarial nets.
\newblock In \emph{NIPS}, 2016.

\bibitem[Duan et~al.(2016)Duan, Chen, Houthooft, Schulman, and
  Abbeel]{duan2016benchmarking}
Yan Duan, Xi~Chen, Rein Houthooft, John Schulman, and Pieter Abbeel.
\newblock Benchmarking deep reinforcement learning for continuous control.
\newblock In \emph{ICML}, 2016.

\bibitem[Ebert et~al.(2018)Ebert, Finn, Dasari, Xie, Lee, and
  Levine]{ebert2018visual}
Frederik Ebert, Chelsea Finn, Sudeep Dasari, Annie Xie, Alex Lee, and Sergey
  Levine.
\newblock Visual foresight: Model-based deep reinforcement learning for
  vision-based robotic control.
\newblock \emph{arXiv preprint arXiv:1812.00568}, 2018.

\bibitem[Essahbi et~al.(2012)Essahbi, Bouzgarrou, and Gogu]{essahbi2012soft}
Nabil Essahbi, Belhassen~Chedli Bouzgarrou, and Grigore Gogu.
\newblock Soft material modeling for robotic manipulation.
\newblock In \emph{Applied Mechanics and Materials}, 2012.

\bibitem[Finn and Levine(2017)]{finn2017deep}
Chelsea Finn and Sergey Levine.
\newblock Deep visual foresight for planning robot motion.
\newblock In \emph{ICRA}, 2017.

\bibitem[Frank et~al.(2011)Frank, Stachniss, Abdo, and
  Burgard]{frank2011efficient}
Barbara Frank, Cyrill Stachniss, Nichola Abdo, and Wolfram Burgard.
\newblock Efficient motion planning for manipulation robots in environments
  with deformable objects.
\newblock In \emph{IROS}, 2011.

\bibitem[Gupta et~al.(2018)Gupta, Murali, Gandhi, and Pinto]{gupta2018robot}
Abhinav Gupta, Adithyavairavan Murali, Dhiraj~Prakashchand Gandhi, and Lerrel
  Pinto.
\newblock Robot learning in homes: Improving generalization and reducing
  dataset bias.
\newblock In \emph{NeurIPS}, 2018.

\bibitem[Haarnoja et~al.(2018)Haarnoja, Zhou, Hartikainen, Tucker, Ha, Tan,
  Kumar, Zhu, Gupta, Abbeel, et~al.]{haarnoja2018soft}
Tuomas Haarnoja, Aurick Zhou, Kristian Hartikainen, George Tucker, Sehoon Ha,
  Jie Tan, Vikash Kumar, Henry Zhu, Abhishek Gupta, Pieter Abbeel, et~al.
\newblock Soft actor-critic algorithms and applications.
\newblock \emph{arXiv preprint}, 2018.

\bibitem[Hafner et~al.(2018)Hafner, Lillicrap, Fischer, Villegas, Ha, Lee, and
  Davidson]{hafner2018learning}
Danijar Hafner, Timothy Lillicrap, Ian Fischer, Ruben Villegas, David Ha,
  Honglak Lee, and James Davidson.
\newblock Learning latent dynamics for planning from pixels.
\newblock \emph{arXiv preprint}, 2018.

\bibitem[Henrich and W{\"o}rn(2012)]{henrich2012robot}
Dominik Henrich and Heinz W{\"o}rn.
\newblock \emph{Robot manipulation of deformable objects}.
\newblock Springer Science \& Business Media, 2012.

\bibitem[Hu et~al.(2018)Hu, Sun, and Pan]{hu2018three}
Zhe Hu, Peigen Sun, and Jia Pan.
\newblock Three-dimensional deformable object manipulation using fast online
  gaussian process regression.
\newblock \emph{RAL}, 2018.

\bibitem[Jia et~al.(2018)Jia, Hu, Pan, Manocha, and Pan]{jia2018learning}
Biao Jia, Zhe Hu, Zherong Pan, Dinesh Manocha, and Jia Pan.
\newblock Learning-based feedback controller for deformable object
  manipulation.
\newblock \emph{arXiv preprint}, 2018.

\bibitem[Jim{\'e}nez(2012)]{jimenez2012survey}
P~Jim{\'e}nez.
\newblock Survey on model-based manipulation planning of deformable objects.
\newblock \emph{Robotics and computer-integrated manufacturing}, 2012.

\bibitem[Kaiser et~al.(2019)Kaiser, Babaeizadeh, Milos, Osinski, Campbell,
  Czechowski, Erhan, Finn, Kozakowski, Levine, et~al.]{kaiser2019model}
Lukasz Kaiser, Mohammad Babaeizadeh, Piotr Milos, Blazej Osinski, Roy~H
  Campbell, Konrad Czechowski, Dumitru Erhan, Chelsea Finn, Piotr Kozakowski,
  Sergey Levine, et~al.
\newblock Model-based reinforcement learning for atari.
\newblock \emph{arXiv preprint}, 2019.

\bibitem[Khalil and Payeur(2010)]{khalil2010dexterous}
Fouad~F Khalil and Pierre Payeur.
\newblock Dexterous robotic manipulation of deformable objects with
  multi-sensory feedback-a review.
\newblock In \emph{Robot Manipulators Trends and Development}. 2010.

\bibitem[Kingma and Ba(2014)]{kingma2014adam}
Diederik Kingma and Jimmy Ba.
\newblock Adam: A method for stochastic optimization.
\newblock \emph{arXiv preprint arXiv:1412.6980}, 2014.

\bibitem[Kumar et~al.(2016)Kumar, Todorov, and Levine]{kumar2016optimal}
Vikash Kumar, Emanuel Todorov, and Sergey Levine.
\newblock Optimal control with learned local models: Application to dexterous
  manipulation.
\newblock In \emph{ICRA}, 2016.

\bibitem[Kurutach et~al.(2018)Kurutach, Tamar, Yang, Russell, and
  Abbeel]{kurutach2018learning}
Thanard Kurutach, Aviv Tamar, Ge~Yang, Stuart~J Russell, and Pieter Abbeel.
\newblock Learning plannable representations with causal infogan.
\newblock In \emph{NeurIPS}, 2018.

\bibitem[Lange and Riedmiller(2010)]{lange2010deep}
Sascha Lange and Martin Riedmiller.
\newblock Deep auto-encoder neural networks in reinforcement learning.
\newblock In \emph{IJCNN}, 2010.

\bibitem[Levine et~al.(2016)Levine, Pastor, Krizhevsky, and
  Quillen]{levine2016learning}
Sergey Levine, Peter Pastor, Alex Krizhevsky, and Deirdre Quillen.
\newblock Learning hand-eye coordination for robotic grasping with deep
  learning and large-scale data collection.
\newblock \emph{ISER}, 2016.

\bibitem[Lillicrap et~al.(2015)Lillicrap, Hunt, Pritzel, Heess, Erez, Tassa,
  Silver, and Wierstra]{lillicrap2015continuous}
Timothy~P Lillicrap, Jonathan~J Hunt, Alexander Pritzel, Nicolas Heess, Tom
  Erez, Yuval Tassa, David Silver, and Daan Wierstra.
\newblock Continuous control with deep reinforcement learning.
\newblock \emph{arXiv preprint}, 2015.

\bibitem[Maas et~al.(2013)Maas, Hannun, and Ng]{maas2013rectifier}
Andrew~L Maas, Awni~Y Hannun, and Andrew~Y Ng.
\newblock Rectifier nonlinearities improve neural network acoustic models.
\newblock In \emph{ICML}, 2013.

\bibitem[Mahler et~al.(2016)Mahler, Pokorny, Hou, Roderick, Laskey, Aubry,
  Kohlhoff, Kröger, Kuffner, and Goldberg]{mahler2016dexnet}
J.~Mahler, F.~T. Pokorny, B.~Hou, M.~Roderick, M.~Laskey, M.~Aubry,
  K.~Kohlhoff, T.~Kröger, J.~Kuffner, and K.~Goldberg.
\newblock Dex-net 1.0: A cloud-based network of 3d objects for robust grasp
  planning using a multi-armed bandit model with correlated rewards.
\newblock In \emph{ICRA}, 2016.

\bibitem[Maitin-Shepard et~al.(2010)Maitin-Shepard, Cusumano-Towner, Lei, and
  Abbeel]{maitin2010cloth}
Jeremy Maitin-Shepard, Marco Cusumano-Towner, Jinna Lei, and Pieter Abbeel.
\newblock Cloth grasp point detection based on multiple-view geometric cues
  with application to robotic towel folding.
\newblock In \emph{ICRA}, 2010.

\bibitem[Matas et~al.(2018)Matas, James, and Davison]{matas2018sim}
Jan Matas, Stephen James, and Andrew~J Davison.
\newblock Sim-to-real reinforcement learning for deformable object
  manipulation.
\newblock \emph{arXiv preprint}, 2018.

\bibitem[McConachie and Berenson(2018)]{mcconachie2018estimating}
Dale McConachie and Dmitry Berenson.
\newblock Estimating model utility for deformable object manipulation using
  multiarmed bandit methods.
\newblock \emph{IEEE Transactions on Automation Science and Engineering}, 2018.

\bibitem[McConachie et~al.(2017)McConachie, Ruan, and
  Berenson]{mcconachie2017interleaving}
Dale McConachie, Mengyao Ruan, and Dmitry Berenson.
\newblock Interleaving planning and control for deformable object manipulation.
\newblock In \emph{ISRR}, 2017.

\bibitem[Mikolov et~al.(2013)Mikolov, Sutskever, Chen, Corrado, and
  Dean]{mikolov2013distributed}
Tomas Mikolov, Ilya Sutskever, Kai Chen, Greg~S Corrado, and Jeff Dean.
\newblock Distributed representations of words and phrases and their
  compositionality.
\newblock In \emph{NIPS}, 2013.

\bibitem[Moll and Kavraki(2006)]{moll2006path}
Mark Moll and Lydia~E Kavraki.
\newblock Path planning for deformable linear objects.
\newblock \emph{T-RO}, 2006.

\bibitem[Nagabandi et~al.(2018)Nagabandi, Kahn, Fearing, and
  Levine]{nagabandi2018neural}
Anusha Nagabandi, Gregory Kahn, Ronald~S Fearing, and Sergey Levine.
\newblock Neural network dynamics for model-based deep reinforcement learning
  with model-free fine-tuning.
\newblock In \emph{ICRA}, 2018.

\bibitem[Nair et~al.(2017)Nair, Chen, Agrawal, Isola, Abbeel, Malik, and
  Levine]{nair2017combining}
Ashvin Nair, Dian Chen, Pulkit Agrawal, Phillip Isola, Pieter Abbeel, Jitendra
  Malik, and Sergey Levine.
\newblock Combining self-supervised learning and imitation for vision-based
  rope manipulation.
\newblock In \emph{ICRA}, 2017.

\bibitem[Oord et~al.(2018)Oord, Li, and Vinyals]{oord2018representation}
Aaron van~den Oord, Yazhe Li, and Oriol Vinyals.
\newblock Representation learning with contrastive predictive coding.
\newblock \emph{arXiv preprint}, 2018.

\bibitem[Piera{\'n}ski et~al.(2001)Piera{\'n}ski, Przyby{\l}, and
  Stasiak]{pieranski2001tight}
Piotr Piera{\'n}ski, Sylwester Przyby{\l}, and Andrzej Stasiak.
\newblock Tight open knots.
\newblock \emph{The European Physical Journal E}, 2001.

\bibitem[Pinto and Gupta(2016)]{pinto2016supersizing}
Lerrel Pinto and Abhinav Gupta.
\newblock Supersizing self-supervision: Learning to grasp from 50k tries and
  700 robot hours.
\newblock \emph{ICRA}, 2016.

\bibitem[Rodriguez et~al.(2006)Rodriguez, Tang, Lien, and
  Amato]{rodriguez2006obstacle}
Samuel Rodriguez, Xinyu Tang, Jyh-Ming Lien, and Nancy~M Amato.
\newblock An obstacle-based rapidly-exploring random tree.
\newblock In \emph{ICRA}, 2006.

\bibitem[Saha and Isto(2007)]{saha2007manipulation}
Mitul Saha and Pekka Isto.
\newblock Manipulation planning for deformable linear objects.
\newblock \emph{T-RO}, 2007.

\bibitem[Schulman et~al.(2013{\natexlab{a}})Schulman, Ho, Lee, and
  Abbeel]{schulman2013generalization}
John Schulman, Jonathan Ho, Cameron Lee, and Pieter Abbeel.
\newblock Generalization in robotic manipulation through the use of non-rigid
  registration.
\newblock In \emph{ISRR}, 2013{\natexlab{a}}.

\bibitem[Schulman et~al.(2013{\natexlab{b}})Schulman, Lee, Ho, and
  Abbeel]{schulman2013tracking}
John Schulman, Alex Lee, Jonathan Ho, and Pieter Abbeel.
\newblock Tracking deformable objects with point clouds.
\newblock In \emph{ICRA}, 2013{\natexlab{b}}.

\bibitem[Schulman et~al.(2015)Schulman, Levine, Abbeel, Jordan, and
  Moritz]{schulman2015trust}
John Schulman, Sergey Levine, Pieter Abbeel, Michael Jordan, and Philipp
  Moritz.
\newblock Trust region policy optimization.
\newblock In \emph{ICML}, 2015.

\bibitem[Seita et~al.(2018)Seita, Jamali, Laskey, Tanwani, Berenstein,
  Baskaran, Iba, Canny, and Goldberg]{seita2018deep}
Daniel Seita, Nawid Jamali, Michael Laskey, Ajay~Kumar Tanwani, Ron Berenstein,
  Prakash Baskaran, Soshi Iba, John Canny, and Ken Goldberg.
\newblock Deep transfer learning of pick points on fabric for robot bed-making.
\newblock \emph{arXiv preprint}, 2018.

\bibitem[Seita et~al.(2019)Seita, Ganapathi, Hoque, Hwang, Cen, Tanwani,
  Balakrishna, Thananjeyan, Ichnowski, Jamali, Yamane, Iba, Canny, and
  Goldberg]{seita2019deep}
Daniel Seita, Aditya Ganapathi, Ryan Hoque, Minho Hwang, Edward Cen, Ajay~Kumar
  Tanwani, Ashwin Balakrishna, Brijen Thananjeyan, Jeffrey Ichnowski, Nawid
  Jamali, Katsu Yamane, Soshi Iba, John Canny, and Ken Goldberg.
\newblock Deep imitation learning of sequential fabric smoothing policies.
\newblock \emph{arXiv preprint}, 2019.

\bibitem[Shimoga(1996)]{shimoga1996robot}
Karun~B Shimoga.
\newblock Robot grasp synthesis algorithms: A survey.
\newblock \emph{IJRR}, 1996.

\bibitem[Smolen and Patriciu(2009)]{smolen2009deformation}
Jerzy Smolen and Alexandru Patriciu.
\newblock Deformation planning for robotic soft tissue manipulation.
\newblock In \emph{Advances in Computer-Human Interactions}, 2009.

\bibitem[Srivastava et~al.(2017)Srivastava, Valkov, Russell, Gutmann, and
  Sutton]{srivastava2017veegan}
Akash Srivastava, Lazar Valkov, Chris Russell, Michael~U Gutmann, and Charles
  Sutton.
\newblock Veegan: Reducing mode collapse in gans using implicit variational
  learning.
\newblock In \emph{NIPS}, 2017.

\bibitem[Stria et~al.(2014)Stria, Prusa, Hlavac, Wagner, Petrik, Krsek, and
  Smutny]{stria2014garment}
Jan Stria, Daniel Prusa, Vaclav Hlavac, Libor Wagner, Vladimir Petrik, Pavel
  Krsek, and Vladimir Smutny.
\newblock Garment perception and its folding using a dual-arm robot.
\newblock In \emph{IROS}, 2014.

\bibitem[Tassa et~al.(2018)Tassa, Doron, Muldal, Erez, Li, Casas, Budden,
  Abdolmaleki, Merel, Lefrancq, et~al.]{tassa2018deepmind}
Yuval Tassa, Yotam Doron, Alistair Muldal, Tom Erez, Yazhe Li, Diego de~Las
  Casas, David Budden, Abbas Abdolmaleki, Josh Merel, Andrew Lefrancq, et~al.
\newblock Deepmind control suite.
\newblock \emph{arXiv preprint}, 2018.

\bibitem[Tian et~al.(2019)Tian, Krishnan, and Isola]{tian2019contrastive}
Yonglong Tian, Dilip Krishnan, and Phillip Isola.
\newblock Contrastive multiview coding.
\newblock \emph{arXiv preprint}, 2019.

\bibitem[Todorov et~al.(2012)Todorov, Erez, and Tassa]{todorov2012mujoco}
Emanuel Todorov, Tom Erez, and Yuval Tassa.
\newblock Mujoco: A physics engine for model-based control.
\newblock In \emph{IROS}, 2012.

\bibitem[Wada et~al.(2001)Wada, Hirai, Kawamura, and Kamiji]{wada2001robust}
Takahiro Wada, Shinichi Hirai, Sadao Kawamura, and Norimasa Kamiji.
\newblock Robust manipulation of deformable objects by a simple pid feedback.
\newblock In \emph{ICRA}, 2001.

\bibitem[Wakamatsu et~al.(2006)Wakamatsu, Arai, and
  Hirai]{wakamatsu2006knotting}
Hidefumi Wakamatsu, Eiji Arai, and Shinichi Hirai.
\newblock Knotting/unknotting manipulation of deformable linear objects.
\newblock \emph{IJRR}, 2006.

\bibitem[Wang et~al.(2019)Wang, Kurutach, Liu, Abbeel, and
  Tamar]{wang2019learning}
Angelina Wang, Thanard Kurutach, Kara Liu, Pieter Abbeel, and Aviv Tamar.
\newblock Learning robotic manipulation through visual planning and acting.
\newblock \emph{RSS}, 2019.

\bibitem[Williams et~al.(2017)Williams, Wagener, Goldfain, Drews, Rehg, Boots,
  and Theodorou]{williams2017information}
Grady Williams, Nolan Wagener, Brian Goldfain, Paul Drews, James~M Rehg, Byron
  Boots, and Evangelos~A Theodorou.
\newblock Information theoretic mpc for model-based reinforcement learning.
\newblock In \emph{ICRA}, 2017.

\bibitem[Wu et~al.(2019)Wu, Yan, Kurutach, Pinto, and Abbeel]{wu2019learning}
Yilin Wu, Wilson Yan, Thanard Kurutach, Lerrel Pinto, and Pieter Abbeel.
\newblock Learning to manipulate deformable objects without demonstrations.
\newblock \emph{arXiv preprint}, 2019.

\bibitem[Yousef et~al.(2011)Yousef, Boukallel, and
  Althoefer]{yousef2011tactile}
Hanna Yousef, Mehdi Boukallel, and Kaspar Althoefer.
\newblock Tactile sensing for dexterous in-hand manipulation in robotics—a
  review.
\newblock \emph{Sensors and Actuators A: physical}, 2011.

\end{thebibliography}


\end{document}